\documentclass[10pt]{article}
\usepackage[letterpaper,margin=1in]{geometry}
\usepackage[T1]{fontenc}
\usepackage{times}
\usepackage{microtype}
\usepackage{graphicx}
\usepackage{caption}
\usepackage{subcaption}
\usepackage{booktabs}
\usepackage{amssymb}
\usepackage{tikz}
\usetikzlibrary{arrows.meta,positioning,fit,calc,shapes.symbols,shapes.geometric}
\newcommand{\rawicoCam}{\tikz[baseline=0.1ex]{\draw[fill=black!85,rounded corners=0.4pt] (0,0) rectangle (0.34,0.22); \draw[fill=white] (0.17,0.11) circle (0.07); \draw[fill=black!85] (0.05,0.22) rectangle (0.13,0.26);}}
\newcommand{\rawicoVial}{\tikz[baseline=0.1ex]{\draw[rounded corners=1pt] (0,0) rectangle (0.12,0.26); \draw[fill=black!85] (-0.015,0.26) rectangle (0.135,0.31); \draw[rounded corners=1pt] (0.18,0) rectangle (0.30,0.26); \draw[fill=black!85] (0.165,0.26) rectangle (0.315,0.31);}}
\newcommand{\rawicoBoard}{\tikz[baseline=0.1ex]{\draw[rounded corners=0.5pt] (0,0) rectangle (0.40,0.26); \draw[fill=black!85] (0.14,0.07) rectangle (0.26,0.19); \foreach \x in {0.05,0.10,...,0.36} {\draw[line width=0.5pt] (\x,0.26)--(\x,0.31);}}}
\newcommand{\rawicoTerm}{\tikz[baseline=0.1ex]{\draw[fill=black!85,rounded corners=0.6pt] (0,0) rectangle (0.36,0.26); \node[text=white,font=\fontsize{5}{5}\selectfont\ttfamily\bfseries,inner sep=0] at (0.18,0.13) {>\_};}}
\newcommand{\rawicoDisk}{\tikz[baseline=0.1ex]{\draw[rounded corners=0.5pt] (0,0) rectangle (0.30,0.26); \draw (0,0.087)--(0.30,0.087); \draw (0,0.173)--(0.30,0.173); \fill (0.24,0.043) circle (0.018); \fill (0.24,0.13) circle (0.018); \fill (0.24,0.216) circle (0.018);}}
\newcommand{\rawicoWeb}{\tikz[baseline=0.1ex]{\draw[rounded corners=0.5pt] (0,0) rectangle (0.36,0.26); \draw[fill=black!85] (0,0.20) rectangle (0.36,0.26);}}
\newcommand{\rawicoGear}{\tikz[baseline=0.1ex]{\draw[line width=0.7pt,-{Latex[length=1.6pt]}] (0.26,0.13) arc (0:300:0.11);}}
\newcommand{\rawicoPulse}{\tikz[baseline=0.1ex]{\draw[line width=0.7pt] (0,0.1)--(0.09,0.1)--(0.13,0.24)--(0.19,0)--(0.24,0.14)--(0.27,0.1)--(0.36,0.1);}}
\newcommand{\rawicoPhone}{\tikz[baseline=0.1ex]{\draw[rounded corners=0.8pt,fill=black!85] (0,0) rectangle (0.16,0.30); \draw[fill=white] (0.02,0.05) rectangle (0.14,0.26);}}
\newcommand{\rawicoCloud}{\tikz[baseline=0.1ex]{\node[cloud,draw,cloud puffs=9,aspect=1.7,minimum width=0.42cm,minimum height=0.26cm,inner sep=0] {};}}
\newsavebox{\boxCam}
\newsavebox{\boxVial}
\newsavebox{\boxBoard}
\newsavebox{\boxTerm}
\newsavebox{\boxDisk}
\newsavebox{\boxWeb}
\newsavebox{\boxGear}
\newsavebox{\boxPulse}
\newsavebox{\boxPhone}
\newsavebox{\boxCloud}
\AtBeginDocument{\sbox{\boxCam}{\rawicoCam}\sbox{\boxVial}{\rawicoVial}\sbox{\boxBoard}{\rawicoBoard}\sbox{\boxTerm}{\rawicoTerm}\sbox{\boxDisk}{\rawicoDisk}\sbox{\boxWeb}{\rawicoWeb}\sbox{\boxGear}{\rawicoGear}\sbox{\boxPulse}{\rawicoPulse}\sbox{\boxPhone}{\rawicoPhone}\sbox{\boxCloud}{\rawicoCloud}}
\newcommand{\icoCam}{\usebox{\boxCam}}
\newcommand{\icoVial}{\usebox{\boxVial}}
\newcommand{\icoBoard}{\usebox{\boxBoard}}
\newcommand{\icoTerm}{\usebox{\boxTerm}}
\newcommand{\icoDisk}{\usebox{\boxDisk}}
\newcommand{\icoWeb}{\usebox{\boxWeb}}
\newcommand{\icoGear}{\usebox{\boxGear}}
\newcommand{\icoPulse}{\usebox{\boxPulse}}
\newcommand{\icoPhone}{\usebox{\boxPhone}}
\newcommand{\icoCloud}{\usebox{\boxCloud}}
\usepackage[colorlinks=true,linkcolor=black,citecolor=black,urlcolor=blue!50!black]{hyperref}
\title{\textbf{Big Brains, Tiny Board:\\Pupavision, an Agentic Physical AI Instrument for\\Monitoring \emph{Drosophila} Pupariation and Lifecycle Transitions}}
\author{Joel Sadler$^{1}$ \quad Daniel Graham$^{1}$ \quad Melanie Worley$^{2}$\\[0.3em]
\small $^{1}$School of Data Science \quad $^{2}$Department of Biology\\
\small University of Virginia\\
\small \texttt{\{jsadler, dgg6b, mworley\}@virginia.edu}}
\date{}

\begin{document}
\maketitle
\vspace{-1.5em}

\begin{abstract}
\noindent Many biology experiments are limited by the availability and ease of use of instruments that monitor and record them. For example, the timing of pupariation in the fruit fly \emph{Drosophila melanogaster}, a widely used model organism, can provide biologically relevant information about growth regulation and is often recorded manually. Here, we present Pupavision, a small physical AI instrument built from a single-board Linux computer and a day/night camera for roughly \$45 to \$75 in parts. The key idea is that a frontier coding agent runs its agentic harness directly on the tiny board. The model thinks in the cloud, but the agent works on the device. The agent assists image capture, writes code to repair errors, and checks the experiment every day to keep raw data capture reliable. If recording stops, the researcher gets a ping on their mobile phone. A simple deterministic capture service records one image every ten minutes and never depends on the agent. We report a 26-day pilot in a fly incubator (about 3,580 frames), including eleven days in which the agent was logged out and capture continued, and a 51-hour outage whose cause led to the layered design. A researcher-in-the-loop workflow then produces per-vial pupariation curves by tracing selected pupae backward in time from a late reference frame. Future work will validate the assay for lifecycle tracking accuracy against expert-labeled data and expand real-time computer vision analysis. Pupavision offers a blueprint for do-it-yourself agentic instruments that put frontier intelligence on a tiny board, with deterministic code underneath. We present this as an early case study from a pilot lab deployment and contribute lessons learned toward a future where scientific domain experts are empowered by DIY agentic hardware.
\end{abstract}

\begin{figure}[h]
\centering
\includegraphics[width=\textwidth]{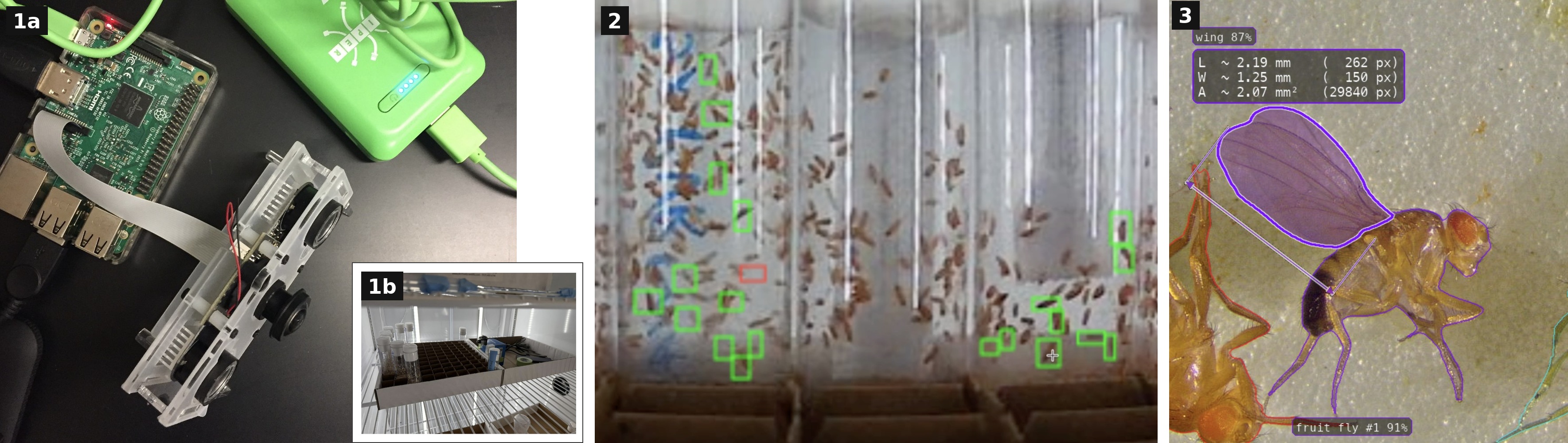}
\caption{Pupavision in three steps. (1a) Build the instrument: a single-board Linux computer with a day/night camera. The agent harness runs on the board. (1b) Place it on a shelf in a temperature-controlled incubator, facing the vials. (2) Capture: the vials are imaged every ten minutes, and the pupae the researcher selects are tracked in the browser interface. (3) After eclosion, the adult flies that emerge can be imaged and analyzed with computer vision for assisted phenotyping, shown here with a segmentation model measuring a wing.}
\label{fig:hero}
\end{figure}

\section{Introduction}
During the life cycle of the fruit fly \emph{Drosophila melanogaster}, pupariation is the transition from the larval to the pupal stage, akin to a caterpillar forming a cocoon. The timing of pupariation is affected by several factors, including nutritional status, body size, the activity of genes that regulate growth, and the activation of regenerative programs (reviewed in \cite{boulan2015,boulan2021}). When internal tissues called imaginal discs are damaged, larvae delay pupariation while the tissues regenerate. The size of that delay gives information about the rate of regeneration and is a standard assay \cite{smithbolton,halme,colombani,worley}. Pupariation is often scored by manual observation, during which a researcher returns to the incubator every few hours, across nights and weekends, and counts the new pupal cases. Some experiments require monitoring every four to six hours, which sets the time resolution of the experiment. The work is laborious and tiring, often because it disrupts sleep, and manual observation therefore limits how many experiments one researcher can run, simultaneously or sequentially.

\begin{figure}[t]
\centering
\begin{tikzpicture}[scale=0.92, transform shape, font=\small, >=Latex,
  box/.style={draw, rounded corners=2pt, minimum height=8mm, align=center, inner sep=4pt},
  inner/.style={draw, rounded corners=2pt, minimum height=9mm, minimum width=52mm, align=center, inner sep=3pt},
  lab/.style={font=\scriptsize, align=center}]
\node[inner, very thick] (cap) at (0,0) {\icoGear\ \textbf{Capture service} (deterministic)\\[-1pt]{\scriptsize one frame every 10 min, restarts itself}};
\node[inner, below=4mm of cap] (store) {\icoDisk\ \textbf{Local storage}\\[-1pt]{\scriptsize original JPEG frames}};
\node[inner, below=4mm of store] (web) {\icoWeb\ \textbf{Web interface}\\[-1pt]{\scriptsize live view, annotator, analysis}};
\node[inner, dashed, below=4mm of web] (agent) {\icoTerm\ \textbf{Agent harness} (Claude Code)\\[-1pt]{\scriptsize runs here, on the board}};
\node[draw, thick, rounded corners=3pt, inner sep=7pt, fit=(cap)(agent), label={[font=\small\bfseries]above:{\icoBoard\ \ Single-board Linux computer}}] (sbc) {};
\node[box, minimum width=24mm, left=30mm of cap] (cam) {\icoCam\\Day/night\\camera};
\node[box, minimum width=24mm, below=4mm of cam] (vials) {\icoVial\\Fly vials};
\node[draw, thick, rounded corners=3pt, inner sep=7pt, fit=(cam)(vials), label={[font=\small\bfseries]above:Incubator}] (inc) {};
\node[box, minimum width=34mm, right=34mm of cap] (hb) {\icoPulse\ Heartbeat monitor\\{\scriptsize (external service)}};
\node[box, minimum width=34mm] (phone) at (hb |- web) {\icoPhone\ Researcher\\{\scriptsize phone or laptop}};
\node[box, minimum width=34mm] (cloud) at (hb |- agent) {\icoCloud\ Cloud model\\{\scriptsize (inference only)}};
\node[draw, thick, rounded corners=3pt, inner sep=7pt, fit=(hb)(cloud), label={[font=\small\bfseries]above:Off the board}] (off) {};
\draw[->, very thick] (cam) -- node[lab, above]{frames} node[lab, below]{USB / CSI} (cap);
\draw[->] (vials) -- (cam);
\draw[->, very thick] (cap) -- (store);
\draw[->] (store) -- (web);
\draw[->] (cap.east) -- node[lab, above]{check-in pings} (hb.west);
\draw[->] (hb) -- node[lab, right, xshift=1pt]{alert if\\pings stop} (phone);
\draw[<->] (web.east) -- node[lab, above]{private mesh} (phone.west);
\draw[<->, dashed] (agent.east) -- node[lab, above]{outbound only} (cloud.west);
\draw[->, dashed] (agent.west) -- ++(-9mm,0) |- node[lab, left, pos=0.25]{writes and\\repairs code} ([yshift=-2.5mm]cap.west);
\draw[->, dashed] ([xshift=-9mm]agent.west |- web.west) -- (web.west);
\end{tikzpicture}
\caption{System composition. The solid path (camera, capture service, storage) is conventional software and needs neither the agent, the network, nor the cloud. The dashed elements are the agent. The harness runs on the board, the model runs in the cloud, and the agent's job is to write, maintain, and check the software around the capture path.}
\label{fig:system}
\end{figure}
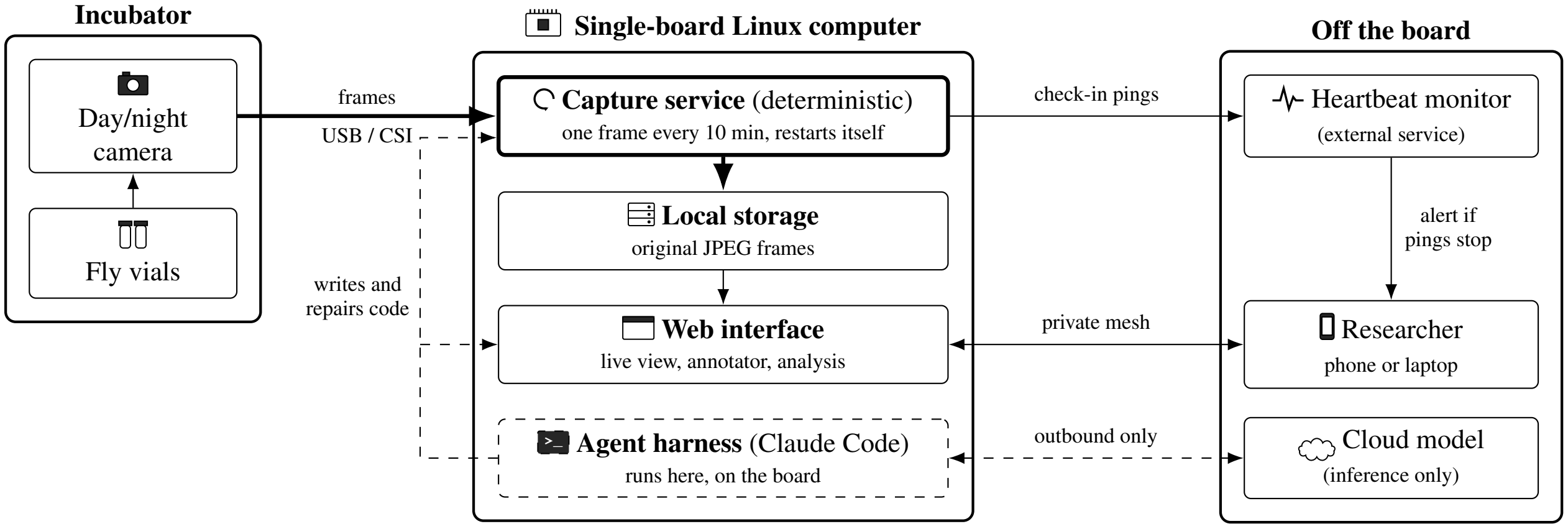

Existing systems automate parts of this. DIAMonDS scores developmental events for individual animals held in multi-well plates on a flatbed scanner, using a trained segmentation model \cite{diamonds}. The Hatching-Box images standard rearing vials with a Raspberry Pi and a fine-tuned object detector that classifies larvae, pupae, and adult flies \cite{hatchingbox}. Both follow a tradition of open, low-cost fly hardware such as ethoscopes \cite{ethoscopes}, and both are built around a trained model and a defined protocol. Pupavision makes a different trade. It has no trained detector, and an agent on the board adapts the capture and analysis code to each lab's setup while the researcher marks the animals of interest. Language-model agents have also been applied to laboratory automation, mostly in expensive settings such as cloud labs, robotic synthesis, and instrument control \cite{boiko,mandal}.

Our starting point is different. For a typical fly lab, the obstacle is not detector accuracy. The obstacle is capturing data reliably around the clock, with as little chance of error as possible. In addition, every experimental setup is different, with different incubators, lighting, vial layouts, and biological questions. Adapting a rig to each setup takes embedded-systems and computer-vision work that most biology labs are not staffed to do. A general-purpose coding agent can now do much of that work from plain-language requests, at the bench. A key human-computer interaction (HCI) insight here is that coding agents can lower a biologist's barrier to entry into physical computing, so that more of their focus goes to experimental creativity and less to embedded-systems code. This builds on a steady trend toward increasingly affordable and powerful \emph{Creative Computing} tools for technical novices, reviewed broadly in \cite{thesis}.

We contribute (1) an architecture in which a coding-agent harness runs on a single-board computer next to the experiment, (2) a resilience pattern in which a deterministic capture service is the floor and the agent is a helper above it, never the only guard, (3) a working readout from a 26-day incubator pilot, and (4) design recommendations for the safe use of agents in instruments of this kind.

\section{System}

The software, setup scripts, and documentation are open source under the MIT license at \url{https://github.com/grassyhilltop/pupavision} \cite{repo}. The repository holds the full technical detail, so this section describes only the design.

\paragraph{Hardware.} We built two versions in two incubators, (1) an Arduino UNO Q running Debian with a 2\,MP USB day/night camera (automatic IR-cut filter, 850\,nm illumination), and (2) a Raspberry Pi with a 5\,MP CSI day/night camera. Each uses wall power, local storage, and a cardboard backdrop. Parts cost roughly \$45 to \$75. Figure~\ref{fig:hero} shows the instrument and Figure~\ref{fig:system} shows how it is composed.

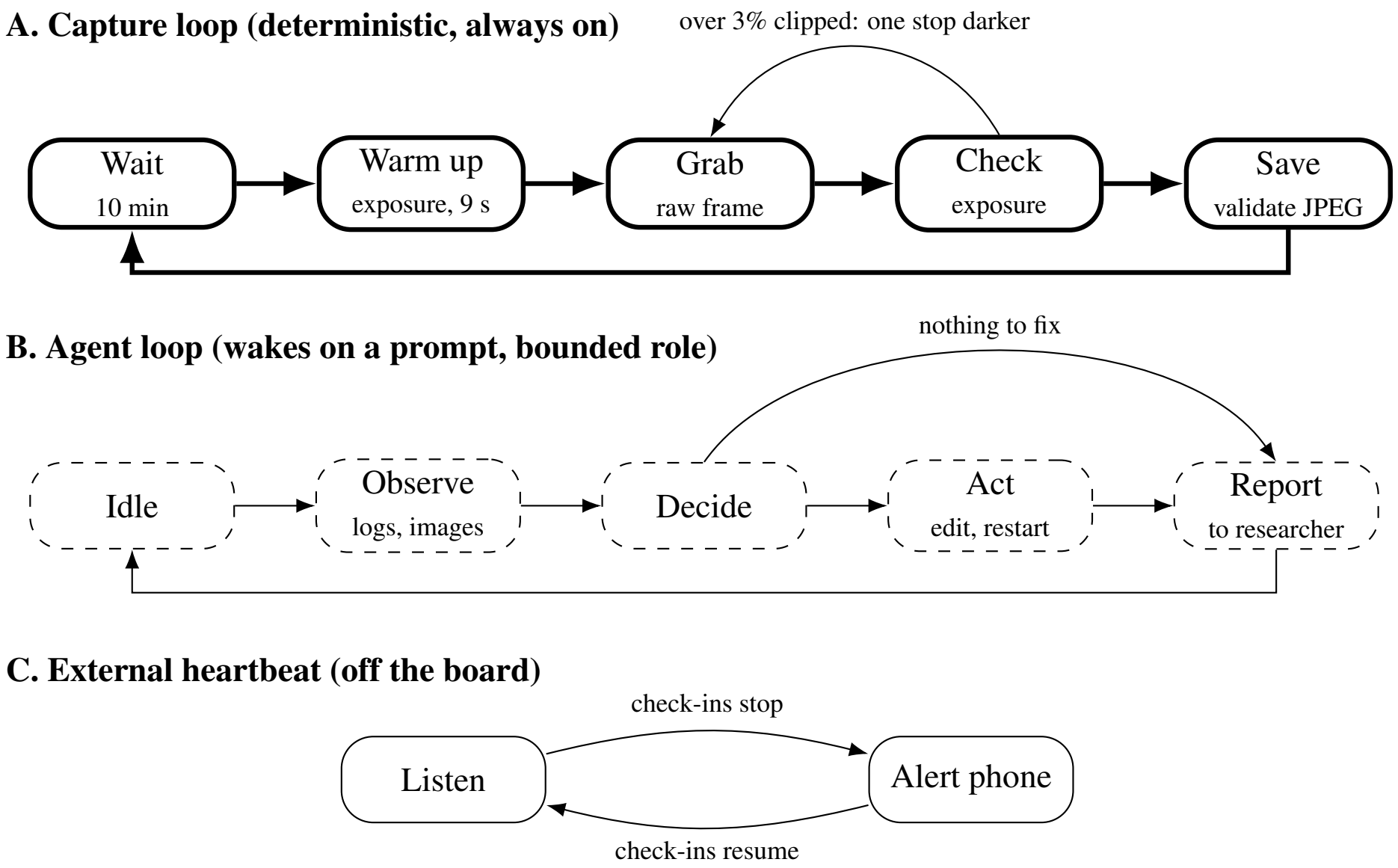
\begin{figure}[t]
\centering
\begin{tikzpicture}[scale=0.95, transform shape, font=\small, >=Latex, node distance=7mm,
  st/.style={draw, rounded corners=8pt, minimum height=8mm, minimum width=19mm, align=center, inner sep=3pt},
  lab/.style={font=\scriptsize, align=center},
  ttl/.style={font=\small\bfseries, anchor=west}]
\node[ttl] at (-1.3,1.45) {A. Capture loop (deterministic, always on)};
\node[st, very thick] (w) at (0,0) {Wait\\{\scriptsize 10 min}};
\node[st, very thick, right=7.5mm of w] (wu) {Warm up\\{\scriptsize exposure, 9 s}};
\node[st, very thick, right=7.5mm of wu] (g) {Grab\\{\scriptsize raw frame}};
\node[st, very thick, right=7.5mm of g] (c) {Check\\{\scriptsize exposure}};
\node[st, very thick, right=7.5mm of c] (s) {Save\\{\scriptsize validate JPEG}};
\draw[->, very thick] (w)--(wu); \draw[->, very thick] (wu)--(g); \draw[->, very thick] (g)--(c); \draw[->, very thick] (c)--(s);
\draw[->, very thick] (s.south) -- ++(0,-4mm) -| (w.south);
\draw[->] (c.north) to[out=120,in=60,looseness=1.2] node[lab, above]{over 3\% clipped: one stop darker} (g.north);
\node[ttl] at (-1.3,-1.55) {B. Agent loop (wakes on a prompt, bounded role)};
\node[st, dashed] (i) at (0,-3.0) {Idle};
\node[st, dashed, right=7.5mm of i] (o) {Observe\\{\scriptsize logs, images}};
\node[st, dashed, right=7.5mm of o] (d) {Decide};
\node[st, dashed, right=7.5mm of d] (a) {Act\\{\scriptsize edit, restart}};
\node[st, dashed, right=7.5mm of a] (r) {Report\\{\scriptsize to researcher}};
\draw[->] (i)--(o); \draw[->] (o)--(d); \draw[->] (d)--(a); \draw[->] (a)--(r);
\draw[->] (r.south) -- ++(0,-4mm) -| (i.south);
\draw[->] (d.north) to[out=55,in=125,looseness=0.8] node[lab, above]{nothing to fix} (r.north);
\node[ttl] at (-1.3,-4.55) {C. External heartbeat (off the board)};
\node[st] (l) at (2.9,-5.55) {Listen};
\node[st, right=30mm of l] (al) {Alert phone};
\draw[->] (l) to[bend left=14] node[lab, above]{check-ins stop} (al);
\draw[->] (al) to[bend left=14] node[lab, below]{check-ins resume} (l);
\end{tikzpicture}
\caption{Logical behavior as three state machines. Loop A is a plain script that produces every measurement, and the operating system restarts it after a crash in any state. Loop B is the agent. It wakes on a daily health check or a message from the researcher, looks, and either repairs something or reports. Loop C receives check-ins from loop A and alerts the researcher when they stop, because neither A nor B can announce its own failure.}
\label{fig:fsm}
\end{figure}

\paragraph{Capture is deterministic.} Capture is its own service under \texttt{systemd}, configured to restart automatically. It targets one $1920\times1080$ frame every ten minutes. Each frame is grabbed uncompressed after about nine seconds of auto-exposure warm-up and encoded as a high-quality JPEG. If more than 3\% of pixels clip, the service grabs again one stop darker. Saved files are validated, and original frames are kept. The service can read the camera directly, so it needs no agent session, no model response, and no internet connection. This is independence from the agent, not a guarantee against power loss, camera failure, or a full disk.

\paragraph{The agent runs on the board.} The Claude Code harness runs on the single-board computer in a persistent terminal session that resumes after a reboot. Model inference is in the cloud, and the board is a thin client that executes tools. The agent can look at images and logs, edit software, and change configuration. Once a day a heartbeat prompt asks it to check uptime, the age of the newest frame, disk space, connectivity, and failed services, and to repair what it can. We chose an existing harness as an engineering convenience and did not compare it with local models or other harnesses. This layer is becoming standardized, first for software tools in the Model Context Protocol \cite{mcp} and now for laboratory hardware in the previewed Model Hardware Standard \cite{mhs}, and we expect instruments like this one to adopt such standards.

\paragraph{An external heartbeat.} An agent cannot report its own death. A separate service receives periodic check-ins from the board and messages the researcher's phone when they stop. Figure~\ref{fig:fsm} shows the three loops and how they relate.

\paragraph{Security boundaries.} The agent needs enough local privilege to repair the instrument unattended, and that same privilege means it could also interrupt capture or change analysis code. We limited exposure in the prototype in four ways. (1) The board is dedicated to this task and holds nothing else. (2) Remote access runs over a private WireGuard mesh with no public inbound ports. (3) Helper services are unprivileged and sandboxed. (4) The board's tool server is read-only by default, with actuation behind an allowlist. These design choices do not guarantee security. They are a combination of best practices that add resilience through a defense-in-depth mindset. Section~\ref{sec:safety} lists the additional security steps we recommend before wider use.

\section{Measuring pupariation with the researcher in the loop}
Fully automatic vial-level detection was attempted and was not reliable enough to trust. What works is a hybrid that spends expertise where it is needed. In a browser annotator that the agent built, the researcher scrubs to a late reference frame in which the pupae have formed and are stationary, draws a box around each target pupa, and labels its vial. Pupae outside the target group, such as those that are morphologically distinct because of a mutation in the \emph{Tubby} gene, are excluded with a click, a judgment that a \emph{Drosophila} biologist makes instantly and that a classifier would need training for. Each vial's denominator is therefore the set of selected, visible pupae, not every pupa in the vial.

\begin{figure}[t]
\centering
\includegraphics[height=1.75in]{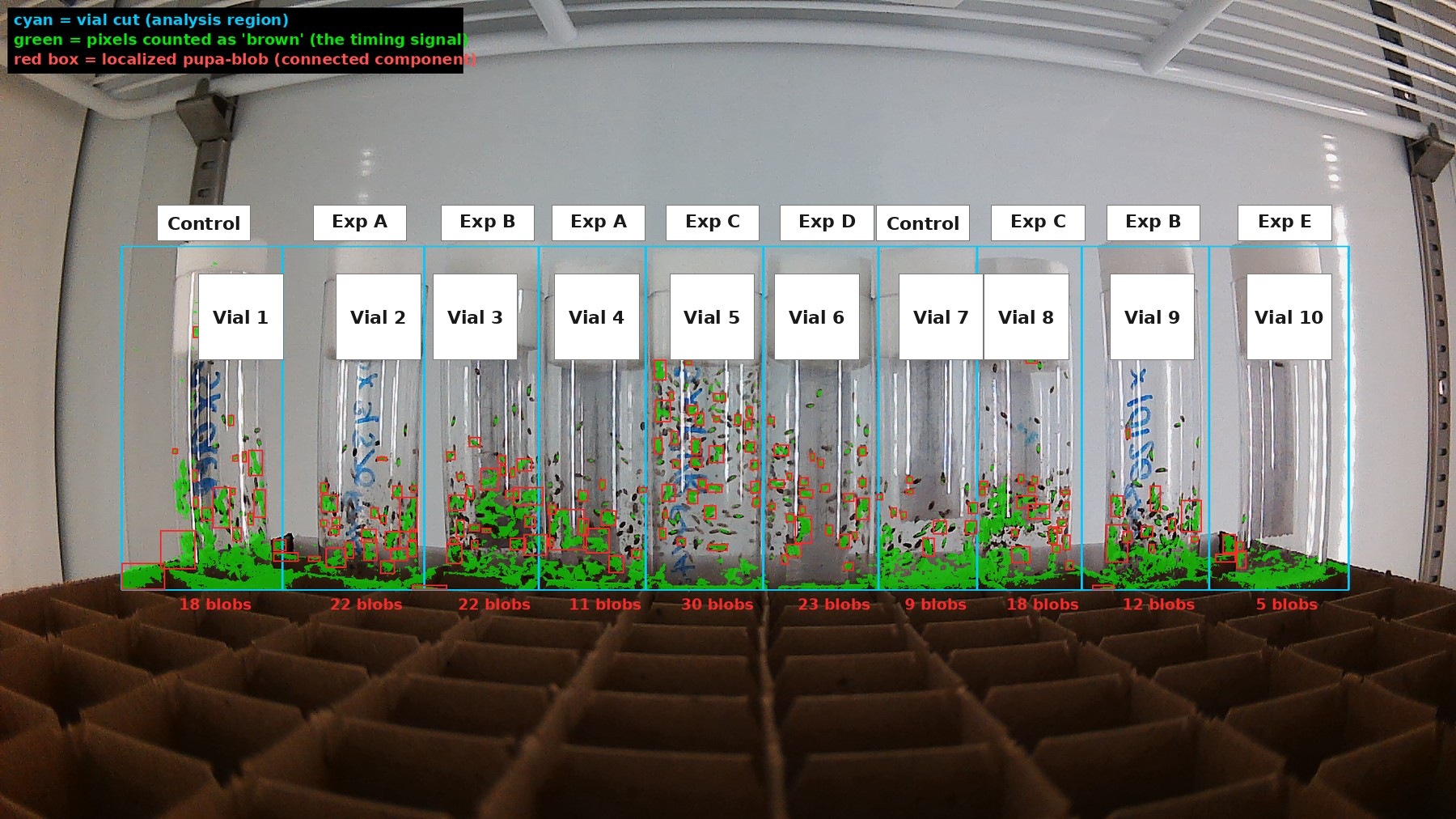}\hfill
\includegraphics[height=1.75in]{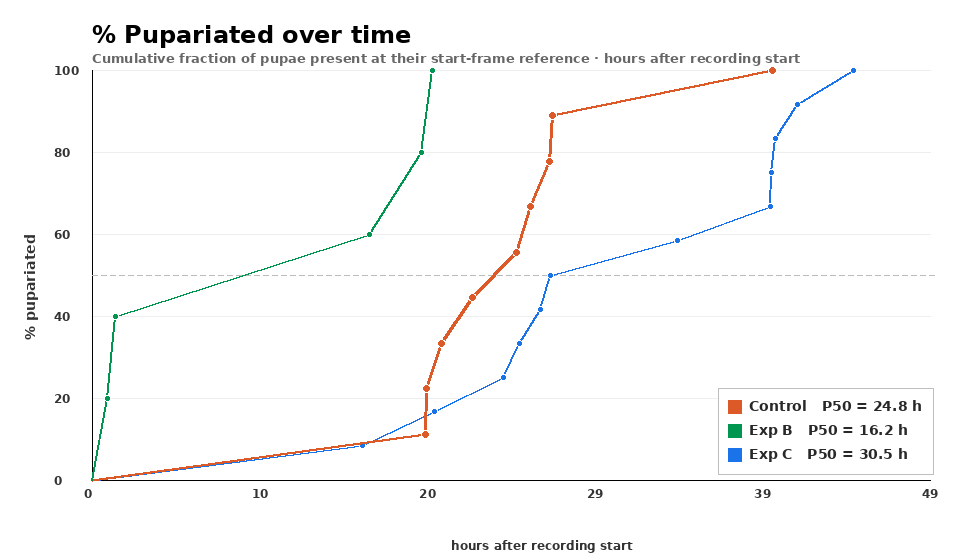}
\caption{Sample data from the rig. Left: a daytime frame with an analysis overlay written on the board by the agent. Vial and group labels are generic. Right: cumulative fraction of selected pupae past the transition for three of the vials, with the median time (P50) for each.}
\label{fig:results}
\end{figure}

Analysis then runs backward in time. At the reference frame each box gets a fixed dark-pixel threshold, and its coverage there is its own reference. Earlier frames are rescaled by a global lighting gain, and a transition is declared at the first frame where coverage falls below 55\% of the reference, with interpolation between frames. Only frames from 08:30 to 20:00 are used, because infrared night frames look different. A transition that happens overnight is therefore placed by interpolation across the gap, so ten-minute capture does not mean ten-minute timing accuracy. The interface draws the cumulative curve beside the boxes so the researcher can check any event against the pixels (Figure~\ref{fig:results}). The threshold crossing is an operational image event. It has not yet been calibrated against expert scoring of puparium formation.

\section{Deployment observations}

\paragraph{The instrument stays up.} One run began on 5 June 2026, in an incubator held at 18\,$^{\circ}$C, and continued for 26 days, storing about 3,580 frames. For eleven of those days the agent was logged out entirely, and capture continued, because capture had never depended on it. The instructive failure came earlier in development. Capture had at one point been an unsupervised loop that pulled frames through the live-stream service. Stopping that service during unrelated dashboard work interrupted acquisition, nothing restarted it, and the series lost 51 hours before anyone noticed. That incident produced the separately supervised capture service, the stall check, and the external heartbeat. The key takeaway is that experiment monitoring cannot be robust if it depends solely on the LLM agent to watch system health, because the agent itself will occasionally be disconnected and unable to tell the researcher that the system is down. The general lesson is to run a parallel safeguard that does not depend on the LLM. The board is expected to check in on a regular heartbeat, and the \emph{absence} of a scheduled ``camera is still operating'' message tells the researcher that something has failed.

\paragraph{The workflow produces a readout.} The pipeline generated per-vial curves (Figure~\ref{fig:results}), with median times to 50\% of 30.5\,h ($n=12$), 24.8\,h ($n=9$), and 16.2\,h ($n=5$) after recording start for the three vials with enough selected pupae. The counts are small because each includes only the pupae the researcher boxed in this first pass, not every pupa in the vial. Pupariation was monitored over about 96 hours, and capture then continued through later stages of the life cycle. These numbers show that the workflow resolves differences between vials. They are not evidence of a genotype effect. One batch of about 257 frames processed on the board in 20 to 30 seconds. Every call is backed by an image that a person can inspect later.

\paragraph{What the agent did.} During development the agent diagnosed dark, low-detail night images, traced them to the camera's compressed video path, and moved capture to uncompressed frames. It added the exposure warm-up, disabled a dynamic frame-rate setting that caused live-view flicker, and suggested a diffuser for an infrared hotspot. It also wrote the annotator, the analysis workflow, and the dashboard. None of this required the biologist to learn camera drivers or image compression. About ten biologists at our institution reviewed successive versions, and their feedback, including a clear preference for drawing boxes over accepting detections they could not verify, shaped the design.

\section{Why put an agent in the loop, and how to do it safely}
\label{sec:safety}
An agent adds non-determinism and risk to an instrument, so it has to earn its place. Here it bought three things. \emph{Accessibility}: the instrument is retargeted in natural language, so a lab without embedded or vision skills can move it between experiments. \emph{Data quality}: the fixes above are ones a non-specialist would not have known to make. \emph{Resilience}: a system that reads its own logs survives long unattended stretches. The instrument is also additive. Manual scoring continues alongside it, so it cannot endanger a running experiment.

Model-level safeguards are not sufficient for a cyber-physical system. An always-on board with local privileges, on a network, in a room with live biology, is a real risk surface, and agents driving instruments have been observed to deviate from instructions \cite{mandal}. Three design choices carried most of the weight in this prototype, and we suggest them as defaults for instruments of this kind: a deterministic fallback that keeps the critical measurement running without the agent, an external heartbeat that tells a person when the system goes quiet, and a human who can physically reach the instrument and pull the plug. Before wider or more autonomous use we recommend three further steps that this prototype does not yet demonstrate: narrowly authorized maintenance commands in place of broad local privilege, protected copies of original images outside the agent's write authority, and reviewed, versioned changes to any code that produces a measurement. Model-level safeguards do not replace operating-system permissions, data protection, and explicit authorization boundaries.

\section{Limitations, future work, and conclusion}
The pupariation values are sample data that show the rig works. They are not a biological finding. The window does not begin at a controlled developmental time zero, night frames are excluded, the vials are a pilot set, and event times have not yet been compared with expert scoring. The next experiment is a scaled run scored in parallel by hand. We also plan to use the same image series to time eclosion, when the adult fly emerges from each pupal case, and to compare movement between genetic backgrounds as a first check for large differences in fitness. We have not compared the agent with a non-agent baseline, and we have not tested whether a non-specialist can deploy the instrument unaided. The biologists we consulted asked most often for higher resolution and for a view of the far side of each vial, which argues for better sensors, mirrors, or more cameras. A single camera is also a single point of failure for the agent's picture of reality.

Two directions follow. The first is real-time overlays: computing useful measures on the live camera feed and drawing them over the image, so that a researcher can adjust an experiment while it runs. This builds on earlier work by one of us on hands-free augmented reality for point-of-view sharing in clinical settings \cite{shluzas}, where the value came from putting the right information in front of the person at the moment of action. The second is scale. At this cost, cameras can be placed wherever an experiment needs watching, which turns a lab into a form of ambient, distributed sensing, with overlays streamed to whichever researcher is responsible.

Broadly, the Pupavision architecture can be applied to a wide range of experiments beyond \emph{Drosophila} pupariation and beyond biology. It fits any experiment that needs continuous visual monitoring, where today a fatigued lab technician or graduate student walks in every few hours to look. The same agentic approach could extend to other model organisms in the life sciences, to materials science and the monitoring of novel material synthesis, to machine monitoring in industrial manufacturing, and even to unconventional uses in computational social science, such as measuring team behavior with ambient sensors, as seen in \cite{kress}. With Pupavision, we have shown an early glimpse of a physical AI future in which scientific domain experts are directly empowered by DIY agentic hardware.

\paragraph{Acknowledgments.} We thank the members of the UVA biology community who gave feedback on early versions. The agent described here wrote much of the instrument's software, and AI tools assisted in drafting and editing this manuscript.

\end{document}